\documentclass[a4paper]{svproc}
\usepackage{bm}
\usepackage{amsfonts}
\usepackage{amsmath}
\usepackage{cleveref}
\usepackage{xspace}
\usepackage[numbers]{natbib}
\usepackage{xcolor}
\usepackage{graphicx}
\usepackage{booktabs}

\usepackage{url}

\newcommand{\solverName}[0]{\emph{Vectorized Online planning wIth Learned diffusion model for POMDP Agents}\xspace}
\newcommand{\solverAbbr}[0]{VOiLA\xspace}

\newcommand{\pomdpTuple}{\ensuremath{\langle\stSpace, \actSpace, \obsSpace, \transF, \obsF, \rewFunc, \gamma\rangle}}

\newcommand{\belSpace}{\ensuremath{\mathcal{B}}\xspace}
\newcommand{\bel}{\ensuremath{b}\xspace}
\newcommand{\belp}{\ensuremath{b'}\xspace}
\newcommand{\belInit}{\ensuremath{b_0}\xspace}
\newcommand{\belS}[1]{\ensuremath{b(#1)}\xspace}

\newcommand{\belTree}{\ensuremath{\mathcal{T}}\xspace}

\newcommand{\belTrans}[1]{\ensuremath{\tau(#1)}\xspace}

\newcommand{\stSpace}{\ensuremath{\mathcal{S}}\xspace}

\newcommand{\st}{\ensuremath{s}\xspace}
\newcommand{\stp}{\ensuremath{s'}\xspace}

\newcommand{\actSpace}{\ensuremath{\mathcal{A}}\xspace}
\newcommand{\act}{\ensuremath{a}\xspace}

\newcommand{\obsSpace}{\ensuremath{\mathcal{O}}\xspace}
\newcommand{\obs}{\ensuremath{o}\xspace}

\newcommand{\transF}{\ensuremath{T}\xspace}
\newcommand{\obsF}{\ensuremath{Z}\xspace}
\newcommand{\transFComp}{\ensuremath{T(\st, \act, \stp)}\xspace}
\newcommand{\obsFComp}{\ensuremath{Z(\stp, \act, \obs)}\xspace}

\newcommand{\rewFunc}{\ensuremath{R}\xspace}
\newcommand{\rewFuncComp}[2]{\ensuremath{R(#1, #2)}\xspace}

\newcommand{\pol}{\ensuremath{\pi}\xspace}
\newcommand{\optPol}{\ensuremath{\pi^*}\xspace}

\newcommand{\valPol}{\ensuremath{V^{\pi}}} 
\newcommand{\valOptPol}{\ensuremath{V^{\pi^*}}} 

\newcommand{\latentNoiseT}{\ensuremath{\omega_T}}
\newcommand{\latentNoiseO}{\ensuremath{\omega_Z}}

\begin{document}

\mainmatter              
\title{VOiLA: Vectorized Online Planning with Learned Diffusion Models for POMDP Agents \vspace{-6pt}}
\titlerunning{VOiLA}  
%
\author{Marcus Hoerger\inst{1} \and Rishikesh Joshi\inst{1} \and Rahul Shome\inst{1}
\and Ian Manchester\inst{2} \and Hanna Kurniawati\inst{1} }
\authorrunning{Marcus Hoerger et al.} 
%
%
\institute{
Australian National University, Canberra, ACT, Australia,\\
\email{\{marcus.hoerger, rishikesh.joshi, rahul.shome, hanna.kurniawati\}@anu.edu.au},\\ 
\and
The University of Sydney, Sydney, NSW, Australia.\\ \email{ian.manchester@sydney.edu.au}}

\maketitle              

\begin{abstract}
\vspace{-12pt}
Planning under uncertainty is an essential capability for autonomous robots. The Partially Observable Markov Decision Process (POMDP) provides a powerful framework for such a capability. Although POMDP-based planning has advanced significantly, its application to real-world problems is often limited by the difficulty of obtaining faithful POMDP models. We present \solverName (\solverAbbr), a framework that learns task-agnostic POMDP world models for online planning under uncertainty. Specifically, \solverAbbr learns transition and observation samplers using conditional diffusion models and observation-likelihood models for particle-based belief updates. To enable efficient online planning, the diffusion samplers are distilled into compact feedforward generators and integrated with Vectorized Online POMDP Planner (VOPP), an online POMDP planner designed to leverage GPU parallelization. Experimental results indicate the distillation strategy reduces sampling cost by up to nearly three orders of magnitude, making learned generative POMDP models practical for online planning. Evaluation of \solverAbbr on three benchmark problems indicates that \solverAbbr achieves equal or better performance than current reinforcement learning (RL) methods for POMDPs, such as Recurrent Soft Actor Critic and DreamerV3, while using less than $10\%$ of the training data, and generalizes much better to unseen environment configurations. In a physical robot experiment, \solverAbbr accomplishes the task in 10 of 10 runs while using models learned purely from simulated data for online planning.
\vspace{-6pt}
\vspace{-12pt}
\end{abstract}
\vspace{-12pt}
\section{Introduction}

\vspace{-6pt}
In the real world, robots must operate reliably despite imprecise actuators, noisy sensors, perception errors, and incomplete knowledge of their surroundings. Partially Observable Markov Decision Processes (POMDPs) provide a principled and unified framework for reasoning under uncertainty, enabling robots to act effectively even when the outcomes of their actions, their own internal state, and the state of the world are not fully known. 

Although solving POMDPs exactly is computationally intractable~\citep{papadimitriou1987complexity}, substantial progress has been made in computing approximate solutions. These methods have largely evolved along two separate but complementary directions. The first is approximate POMDP solving~\citep{kurniawati22Partially}, which today is often performed online and therefore avoids substantial offline pre-computation, but requires POMDP models to be available a priori. The second is reinforcement-learning-based approaches, starting with~\citep{hausknecht2015deep}, which  relax the need for explicit models but generally require large amounts of data and significant training time. 

Methods have been proposed to combine the strengths of planning and learning. Some~\citep{deglurkar2023compositional,lee.rss,moss2024betazero,schrittwieser2020mastering} use search tree structures typical of planning and integrate them with learning of certain components of the POMDP models, heuristics, or estimates of the value functions. However, the planning components in these combinations are single-threaded. We adopt this integrated planning-and-learning paradigm, but build on recent learning techniques and a recent online POMDP planner compatible with modern learning platforms.

Specifically, we propose an approximate POMDP solving framework \solverName (\solverAbbr). In POMDP planning, the required world models include the transition models, which capture the uncertainty of action outcomes as a conditional probability distribution function over possible next states, and observation models, which capture the likelihood of observations under different pairs of states and actions. \solverAbbr models the POMDP transition and observation functions using expressive conditional diffusion models. To enable fast sampling, these models are distilled into compact feedforward generators. In addition, \solverAbbr learns a contrastive likelihood model for belief updates and online planning. All models are learned offline purely from simulated data. 

Once learned, the models are used by VOPP~\citep{hoerger2026vectorizedonlinepomdpplanning} to compute near-optimal POMDP policies online. VOPP is a recent online POMDP planner that vectorizes planning computations to exploit the massive parallelism of GPUs. For \solverAbbr, we extend VOPP to handle continuous observation spaces. By learning only the transition and observation models and then using them for planning, \solverAbbr can significantly reduce data requirements while improving generalization across tasks and scenarios. 

We evaluate \solverAbbr on three domains: a simple POMDP benchmark introduced in~\citep{deglurkar2023compositional}, a simulated quadruped locomotion task in uncertain and previously unseen environments, and a physical robot experiment on target finding in an uncertain environment using a quadruped. 
The simulation results indicate that \solverAbbr achieves comparable performance to current RL-based POMDP methods, DreamerV3~\citep{hafner2025mastering} and Recurrent Soft Actor Critic for Discrete Actions~\citep{ni2022recurrent},
in trained scenarios while using only $10\%$ of their training data, and exhibits stronger generalization to unseen scenarios.
In a physical target finding problem involving a quadruped robot, \solverAbbr uses models learned purely from simulated data for online planning and successfully accomplishes the task in 10 of 10 runs. This indicates the potential that models learned solely from simulation data can be directly used to generate good real-world strategies under the POMDP framework.

\vspace{-0.5cm}
\section{Background \& Related Work}

\subsection{Partially Observable Markov Decision Process (POMDPs)}

A POMDP problem is specified as a 7-tuple $\pomdpTuple$, where $\stSpace$, $\actSpace$, and $\obsSpace$ denote the state, action, and observation spaces. In this paper, $\stSpace$ and $\obsSpace$ may be discrete, continuous, or hybrid, while $\actSpace$ is discrete; for simplicity, we describe the background assuming all spaces are discrete. The transition function $\transF$ is a conditional distribution, with $\transFComp = P(\stp | \st, \act)$ denoting the probability of next state $\stp$ after performing action $\act$ in state $\st$. The observation function $\obsF$ is defined by $\obsFComp = P(\obs | \stp, \act)$, and the reward function $\rewFunc: \stSpace \times \actSpace \rightarrow \mathbb{R}$ is bounded, and $\gamma \in (0, 1)$ is a discount factor.

Since the agent does not directly observe the true state, it maintains a belief $\bel \in \belSpace$, i.e., a probability distribution over states, where $\belSpace$ is the set of all possible beliefs. In almost all scalable approximate POMDP solvers, the agent starts from a given initial belief $\belInit \in \belSpace$. Given belief $\bel$, action $\act$, and observation $\obs$, the belief is updated according to $\belp{(\stp)} = \belTrans{\bel, \act, \obs} \propto \obsFComp \sum_{s \in \stSpace} \transFComp \bel(s)$, where $\belTrans{\cdot}$ is the belief update operator. The expected immediate reward is $\rewFuncComp{\bel}{\act} = \sum_{s \in \stSpace} \rewFuncComp{\st}{\act} \belS{\st}$.

The actions a POMDP agent takes at each time step is governed by a policy $\pol: \belSpace \rightarrow \actSpace$, which maps beliefs  to actions. In this paper, we assume a deterministic policy, where a belief is mapped to an action, rather than a distribution over the action space. Since beliefs are sufficient statistics of action-observation histories, this is equivalent to mapping histories to actions. Each policy $\pol$ has value $\valPol: \belSpace \rightarrow \mathbb{R}$, defined as the expected discounted return $\valPol(\bel) := E[\sum_{t = 0}^\infty \gamma^t \rewFuncComp{\bel}{\act}]$, where the expectation is taken over trajectories induced by the policy, transition function, and observation function. The POMDP solution is an optimal policy $\optPol$ with value $\valOptPol(\bel) = \sup_{\forall \pol} \valPol(\bel)$ for all $\bel \in \belSpace$.

\vspace{-6pt}
\subsection{Online POMDP Planning}\label{ssec:online_pomdp_solvers}
Online POMDP planning interleave policy computation and policy execution by maintaining a belief and performing forward search from the current belief. Earlier approaches such as POMCP~\citep{silver2010montee} use Monte Carlo Tree Search combined with particle beliefs to scale to large state spaces. Subsequent methods, including DESPOT~\citep{Ye2017despot} and ABT~\citep{kurniawati2016online}, improve efficiency through scenario-based planning and belief-tree reuse. More recent extensions such as POMCPOW and PFT-DPW~\citep{sunberg2018online} address continuous observation spaces using progressive widening. While these methods have demonstrated strong planning performance, they were primarily designed around serial CPU-based simulation and do not fully exploit modern parallel hardware such as GPUs. Although learned transition and observation models can be integrated into these solvers, their architectures do not efficiently leverage batched neural-network inference during online planning.

Vectorized Online POMDP Planner (VOPP)~\citep{hoerger2026vectorizedonlinepomdpplanning} is a recently proposed massively parallel online method based on the Partially Observable Reference Policy Programming (PORPP) framework~\citep{kim2025porpp}, which partially solves the POMDP value function analytically, leaving the numerical computation to be only estimation of expected total reward: 
$V(b) = \frac{1}{\eta}
\log
\left(
\sum_{a \in \mathcal{A}}
\exp(\eta \Psi(b,a))
\right):= [\mathcal{L}_{\eta}\Psi](\bel)$, where $\eta > 0$ is a temperature parameter, balancing reward maximization with deviation from the reference policy, and $\mathcal{L}_{\eta}$ is the log-sum-exp operator~\cite{blanchard2021accurately}. The expression $\Psi$ denotes {\em preferences} over belief-action pairs
\begin{equation}
\vspace{-3pt}
\Psi(b,a) = \frac{1}{\eta}\log(\pol_0(\act \mid \bel))+ \rewFunc(\bel, \act)+ \gamma \sum_{\obs\in\obsSpace} \obsF(\bel, \act, \obs)[\mathcal{L}_{\eta}\Psi](\tau(\bel,\act,\obs)),
\vspace{-3pt}
\end{equation}
where $\rewFunc(\bel, \act)$ is a Monte Carlo estimate of $\int_{\st\in\stSpace}R(s, a)\bel(s)\mathrm{d}s$. By removing the need to interleave numerical optimisation and estimation of expectation, the scheduling to parallelise POMDP solving can be substantially reduced.  

Building on this formulation, VOPP represents the entire belief tree using a collection of tensors storing belief nodes, action nodes, and action preferences. It incrementally constructs the belief tree by interleaving fully vectorized forward search using tens of thousands of parallel simulations with fully vectorized preference backup operations. Both operations are implemented entirely as tensor computations running on the GPU, enabling VOPP to achieve state-of-the-art performance in online POMDP planning.

\vspace{-12pt}
\subsection{Learning-Based Approaches in POMDPs}
Learning approaches have been proposed to compute POMDP policies. One typical approach is Recurrent model-free Reinforcement Learning. They 
augment policies with memory mechanisms such as recurrent neural networks~\citep{hausknecht2015deep,kapturowski2018recurrent,ni2022recurrent} or transformers~\citep{parisotto2020stabilizing}, allowing them to summarize action-observation histories without explicitly maintaining a belief state. Model-based Reinforcement Learning methods \cite{lee2020stochastic} and end-to-end learning \cite{Col22:lcinet,karkus2017qmdp} for approximate POMDP solving have been proposed too. While effective on the training distribution, such approaches typically require large amounts of task-specific training data.

An alternative is to learn models for planning. Methods such as MuZero~\citep{schrittwieser2020mastering}, Dreamer~\citep{hafner2025mastering}, and Visual Tree Search (VTS)~\citep{deglurkar2023compositional} learn dynamics and/or observation models that are used for planning, policy optimization, or belief updates. However, these methods primarily target task-specific learning and are therefore tightly coupled to the training objectives and environments used during learning.

In contrast, \solverAbbr learns task-agnostic transition and observation models designed specifically for integration with a scalable online POMDP planner. The learned models can therefore be reused across tasks sharing the same environment dynamics while supporting both online planning and particle-based belief updates. The highly scalable POMDP planner of \solverAbbr enables good policies to be inferred despite model errors, further reducing the data requirements. 

\vspace{-6pt}
\section{Overview of \solverAbbr}\label{sec:overview}
\vspace{-6pt}


\begin{figure*}[t]
\centering
\includegraphics[width=\textwidth]{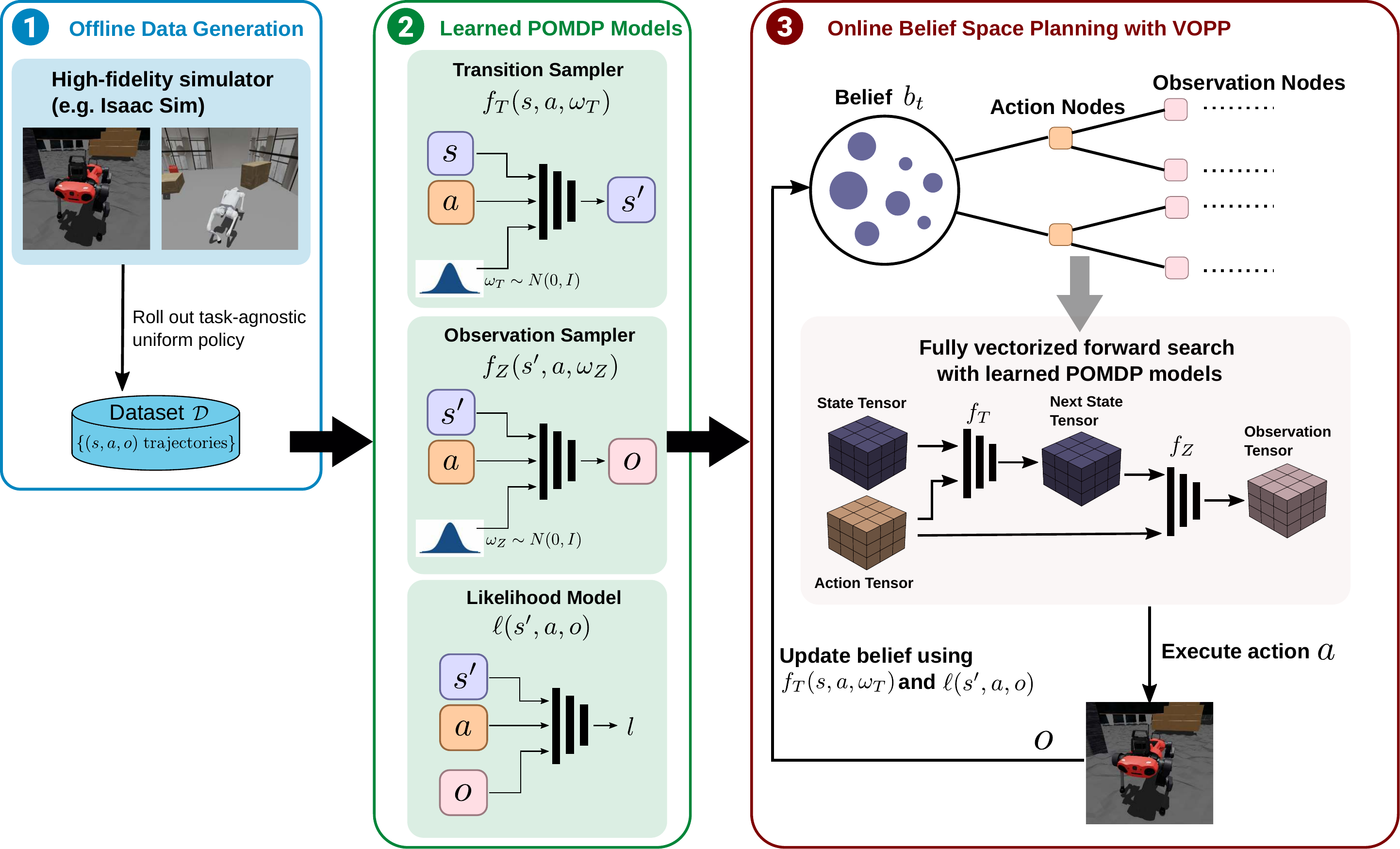}
\caption{Overview of \solverAbbr. Offline trajectories generated in a high-fidelity simulator using a task-agnostic policy are used to learn a transition sampler $f_{\transF}$, an observation sampler $f_{\obsF}$, and a likelihood model $\ell$. During online planning, these models are integrated into the vectorized POMDP planner VOPP, which performs belief-space planning through thousands of parallel simulations and belief updates.}
\label{fig:method}
\vspace{-0.5cm}
\end{figure*}

We propose \solverAbbr, a sequential decision-making framework that combines the recently proposed massively parallel online POMDP planning VOPP with learned neural-network parameterized POMDP models for scalable planning under uncertainty. \solverAbbr learns task-agnostic POMDP models purely from offline-generated simulation data and uses them with VOPP to enable efficient online belief-space planning. An overview of \solverAbbr is shown in \Cref{fig:method}.

In particular, our method learns three POMDP model components: (a) a transition sampler $f_\transF: \stSpace \times \actSpace \mapsto \stSpace$ to approximately sample next states according to the transition function \transFComp, (b) an observation sampler $f_\obsF: \stSpace \times \actSpace \mapsto \obsSpace$ that approximately samples observations according to \obsFComp, where high-dimensional observations are modeled in a learned latent observation space, and (c) a likelihood model $\ell: \stSpace\times\actSpace\times\obsSpace \mapsto \mathbb{R}$ which approximates the (unnormalized) POMDP observation density \obsF.

A core requirement of the learned transition and observation samplers is that they must be sufficiently expressive to capture complex stochastic robot dynamics (e.g., legged locomotion) and high-dimensional observation distributions (e.g., images), while remaining efficient enough to serve as generative models during online POMDP planning. We achieve this by training both the transition and observation samplers in two stages. First, expressive conditional diffusion models are trained to model the underlying transition and observation distributions. Second, the diffusion samplers are distilled into compact feedforward generators that closely approximate the original diffusion models while reducing sampling cost by up to nearly three orders of magnitude. Details on our model learning approach are provided in \Cref{ssec:pomdp_model_learning}.

The learned transition, observation, and likelihood models are integrated into VOPP for both online belief-space planning and particle-based belief updates. During planning, vectorized forward simulations are generated directly using the offline learned transition and observation samplers instead of relying on hand-crafted models. To support continuous observation spaces, we additionally extend VOPP with a fully vectorized implementation of progressive widening~\citep{sunberg2018online}. 

After computing an action for the current belief, \solverAbbr executes the action in the environment and updates the belief based on the perceived observation. Beliefs are represented as sets of particles and updated using a Sequential Importance Resampling (SIR) particle filter~\citep{arulampalam2002tutorial}. The transition sampler $f_\transF$ is used to sample proposal particles, while the likelihood model $\ell$ is used to compute particle weights from observations before resampling. Details on how the learned models are integrated into VOPP are provided in \Cref{ssec:online_planning}.

\vspace{-6pt}
\section{Details of \solverAbbr}
\vspace{-6pt}
\subsection{Offline Data Generation}
\label{ssec:offline_data_generation}

Training data for the learned POMDP models is generated entirely offline using high-fidelity robot simulators such as Isaac Sim~\citep{mittal2025isaac}. To generate training data, we execute a task-independent data collection policy (a uniform policy in all experiments) in simulated environments and record trajectories consisting of POMDP states, actions, and observations.

Formally, we construct an offline dataset $\mathcal{D} = \{\xi^{(n)}\}_{n=1}^{N}$, where each trajectory $\xi^{(n)} = \{(s_i^{(n)}, a_i^{(n)}, o_i^{(n)})\}_{i=1}^{K}$ consists of a sequence of states, actions, and observations generated through interaction with the simulator. Here, $N$ denotes the number of trajectories and $K$ their maximum length.

Because the data collection policy is independent of the downstream task and reward structure, the resulting dataset captures general dynamics and observation structure rather than task-specific behaviors. The dataset $\mathcal{D}$ is then used to train the transition sampler, observation sampler, and likelihood model.

\vspace{-6pt}
\subsection{POMDP Model Learning Approach}
\label{ssec:pomdp_model_learning}

Learning POMDP models for real-world robotic planning under uncertainty is challenging due to complex stochastic dynamics, high-dimensional observations, and the computational requirements of online belief-space planning. The learned models must not only accurately represent the underlying transition and observation distributions, but also support efficient sampling during online planning.

\solverAbbr learns three efficient model components for online planning: a transition sampler $f_\transF$ that approximately samples successor states according to the transition distribution \transFComp, an observation sampler $f_\obsF$ that approximately samples observations according to the observation distribution $\obsF(\stp,\act,\obs)$, and a likelihood model $\ell(\stp, \act, \obs)$ that approximates the (unnormalized) observation density $\obsF$. Details on how these model components are learned, including training objectives and derivations, are presented in the following subsections.

\vspace{-6pt}
\subsubsection{Transition Sampler}\label{ssec:transition_sampler}

The transition sampler is modeled as
\begin{equation}
    \stp = f_\transF(\st,\act,\latentNoiseT),
\end{equation}
where $\latentNoiseT \sim \mathcal{N}(0,I)$ is a latent noise variable that captures transition stochasticity. Real-world transition dynamics are often highly nonlinear and multimodal due to contact dynamics, actuation errors, and robot--environment interactions. Capturing such dynamics therefore requires expressive generative models.

Motivated by their strong performance in modeling complex multimodal distributions, we first train a conditional denoising diffusion model~\citep{ho2020denoising} on transition tuples $(\st,\act,\stp)\sim \mathcal{D}$ using a v-prediction objective~\citep{salimans2022progressive}. In particular, we construct a noise-corrupted next state $x_\tau = \sqrt{\bar{\alpha}_{\tau}}\stp + \sqrt{1-\bar{\alpha}_{\tau}}\epsilon,$ where $\epsilon \sim \mathcal{N}(0,I)$, $\tau \in (0,1)$ denotes the diffusion time, and $\bar{\alpha}_{\tau} \in (0, 1)$ is the cumulative noise schedule (we use a cosine schedule) evaluated at $\tau$. A conditional denoiser $v_\transF(\st,\act,x_\tau,\tau)$ is then trained to predict the v-target $v = \sqrt{\bar{\alpha}_\tau}\epsilon - \sqrt{1-\bar{\alpha}_\tau}\stp$.

While diffusion models are highly expressive, their iterative sampling process is computationally too expensive for online POMDP planning, where a large number of forward simulations must be generated.

To enable efficient online planning, we distill an ODE-based diffusion sampler~\citep{song2021denoising} into the compact model $f_\transF$, which directly maps $(\st,\act,\latentNoiseT)$ to a next-state sample \stp, such that \stp is approximately distributed according to $\transF(\st, \act, \stp)$. We train $f_\transF$ to imitate the next-state samples generated by the ODE-based diffusion sampler. Specifically, we minimize the distillation loss
\begin{equation}
\mathcal{L}_{\mathrm{dist}}
=
\mathbb{E}_{(\st,\act,\stp)\sim\mathcal{D},\,\latentNoiseT\sim\mathcal{N}(0,I)}
\left[
\left\|
\hat{\stp} - f_\transF(\st,\act,\latentNoiseT)
\right\|_2^2
\right],
\end{equation}
where $\hat{\stp}$ denotes the next-state sample generated by the ODE-based diffusion sampler using $(\st,\act,\latentNoiseT)$ and denoiser $v_\transF$.

The transition sampler $f_\transF$ closely approximates the ODE-based sampler, while reducing sampling cost by up to nearly three orders of magnitude.

\vspace{-6pt}
\subsubsection{Observation Sampler}\label{ssec:observation_sampler}

The observation sampler $f_\obsF$ is learned analogously to the transition sampler described in the previous paragraph. In particular, $f_\obsF$ is a compact distilled sampler that samples observations conditioned on $(\stp,\act)$ for efficient use during online planning. For low-dimensional observations, the observation sampler generates observations in the original observation space $\mathcal{O}$ directly. In this case, the same diffusion and distillation learning procedure described for the transition sampler is used to train $f_\obsF$.

However, for high-dimensional observations such as images, sampling observations from the original observation space $\mathcal{O}$ can be computationally inefficient during planning. In such cases, we instead introduce a lower-dimensional latent observation space $\mathcal{Z}$ with $\dim(\mathcal{Z}) \ll \dim(\mathcal{O})$ using a conditional observation autoencoder consisting of an encoder $E$ and decoder $D$, such that $z = E(\stp,\act,\obs)$ and $\obs = D(\stp,\act,z)$. The encoder and decoder are trained jointly on tuples $(\stp,\act,\obs)\sim\mathcal{D}$ by minimizing the reconstruction objective
\begin{equation}
\mathcal{L}_{\mathrm{rec}}
=
\mathbb{E}_{(\stp,\act,\obs)\sim\mathcal{D}}
\left[
\left\|
\obs - D(\stp,\act,E(\stp,\act,\obs))
\right\|_2^2
\right].
\end{equation}

Different encoder and decoder architectures can be used depending on the observation modality, including vector-valued, image-based, or mixed observations. Once trained, the autoencoder weights are frozen and the encoder $E$ is used to map observations to the latent observation space $\mathcal{Z}$.

We then train a conditional diffusion model in the latent observation space using tuples $(\stp,\act,z)$, where $z = E(\stp,\act,\obs)$. Following the same diffusion and distillation procedure used for the transition sampler, we finally distill the diffusion sampler into the compact feedforward observation sampler
\begin{equation}
z = f_\obsF(\stp,\act,\latentNoiseO),
\end{equation}
where $\latentNoiseO \sim \mathcal{N}(0,I)$ is latent noise. This enables efficient latent observation sampling during online planning.




\vspace{-6pt}
\subsubsection{Likelihood Model}\label{ssec:likelihood_model}

Online belief updates and planning with continuous observation spaces require evaluating the observation density $\obsF(\stp,\act,\obs)$ for candidate next states. However, learning normalized observation densities is challenging in practice due to the intractability of the normalization constants~\citep{lecun2006tutorial}. Fortunately, particle-based belief updates only require relative particle weights, making unnormalized observation densities sufficient for online POMDP planning.

We therefore learn an unnormalized likelihood model $\ell(\stp,\act,\obs)$ whose exponentiated outputs approximate the observation density $\obsF(\stp,\act,\obs)$ up to a normalizing constant. The model $\ell$ is trained contrastively using an InfoNCE-style objective~\citep{oord2018representation}: for a mini-batch of size $B$, i.e., $\{(\stp_i,\act_i,\obs_i)\}_{i=1}^B \sim \mathcal{D}$, the model is evaluated on all $B \times B$ state-action--observation combinations, where matching tuples $(\stp_i,\act_i,\obs_i)$ are treated as positive pairs and mismatched tuples $(\stp_i,\act_i,\obs_j)$ with $i \neq j$ serve as in-batch negatives. We then minimize the InfoNCE loss
\begin{equation}
    \mathcal{L}_{\mathrm{likelihood}}
    =
    -\frac{1}{B}
    \sum_{i=1}^{B}
    \log
    \frac{
        \exp(\ell(\stp_i, \act_i, \obs_i))
    }{
        \sum_{j=1}^{B}\exp(\ell(\stp_i, \act_i, \obs_j))
    }.
\end{equation}
This objective encourages the likelihood model to assign higher values to matching $(\stp,\act,\obs)$ tuples from the dataset than to mismatched tuples.

The resulting model $\ell$ does not represent a normalized observation density. Instead, exponentiated likelihood values $\exp(\ell(\stp,\act,\obs)/\lambda)$ are used as relative particle weights during belief updates, which is sufficient for particle-based online POMDP planning. The parameter $\lambda > 0$ is a problem-dependent temperature parameter which controls the sharpness of the resulting particle weight distribution, with smaller values producing more concentrated particle weights and larger values yielding more uniform particle weights, thereby mitigating particle impoverishment during belief updates. In our experiments in \Cref{sec:experiments}, we use $\lambda \in \{1.0, 1.25\}$, which we found to perform well.

\vspace{-9pt}
\subsection{Online belief-space planning with \solverAbbr}\label{ssec:online_planning}
During online planning, \solverAbbr follows VOPP's vectorized belief-space planning procedure. In the original VOPP algorithm, the belief tree \belTree is expanded using a hand-crafted vectorized generative model $G(\st,\act)$ that samples successor states and observations during forward search. In \solverAbbr, we instead replace this generative model with the learned transition and observation samplers introduced in \Cref{ssec:pomdp_model_learning}. Given batched states and actions, the transition sampler $f_\transF(\st,\act)$ generates successor states $\stp$, while the observation sampler $f_\obsF(\stp,\act)$ samples corresponding observations $\obs$.

To handle continuous observation spaces, we extend VOPP with a vectorized implementation of Progressive Widening (PW)~\citep{sunberg2018online}. PW  incrementally limits the number of observation branches associated with each action node in \belTree. In particular, an action node $a$ is expanded with a sampled observation whenever $|C(\act)| < k_\obs N(\act)^{\alpha_\obs}$, where $|C(\act)|$ denotes the number of child observation branches associated with action node $\act$, $N(\act)$ is the action-node visitation count, and $k_\obs,\alpha_\obs > 0$ are widening hyperparameters. When the widening criterion is satisfied at action node \act, a newly sampled observation is added as a new observation branch by extending the observation nodes tensor data structure. Otherwise, the sampled successor state is associated with a previously existing observation branch of the same action node. The likelihood model $\ell(\stp,\act,\obs)$ is then used to weight successor states according to the observation stored at the selected branch, followed by resampling.

Importantly, parallel simulations are grouped by their parent action nodes, after which the above PW operations are performed for all action-node groups in parallel. This preserves VOPP's massively parallel planning formulation, even in continuous observation spaces.


\vspace{-12pt}
\section{Experiments \& Results}
\label{sec:experiments}
\vspace{-6pt}

We tested \solverAbbr on three planning under uncertainty problems, detailed below. 

\vspace{-9pt}
\subsection{Problem Scenarios}\label{ssec:scenarios}
\begin{figure*}[t]
\centering
\begin{tabular}{ccc}
    \includegraphics[height=2.3cm]{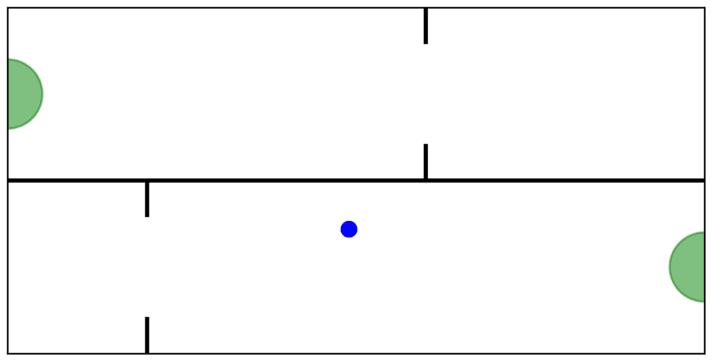} &
    \includegraphics[height=2.3cm]{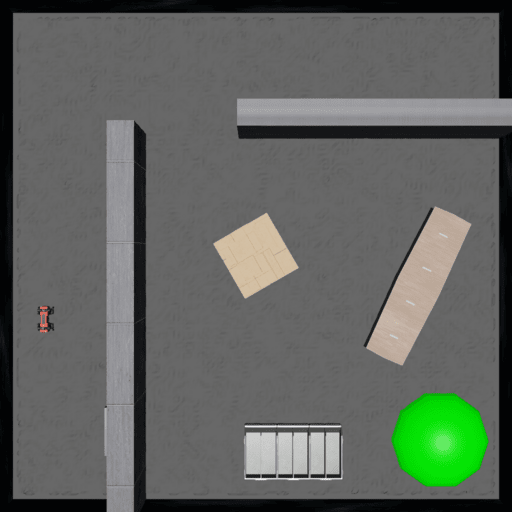} &
    \includegraphics[height=2.3cm]{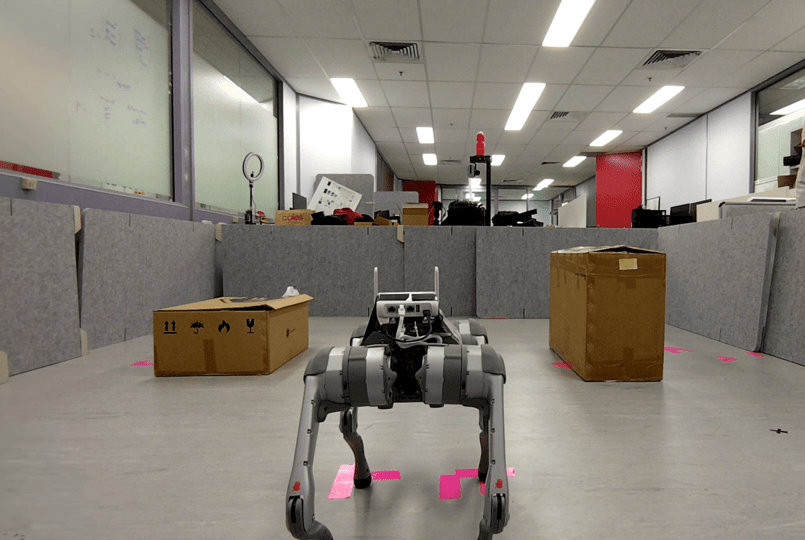} \\
    \small (a) \textsc{FloorPositioning} &
    \small (b) \textsc{LeggedNavigation} &
    \small (c) \textsc{TargetFinding}
\end{tabular}

\caption{Problem scenarios used to evaluate \solverAbbr.}
\label{fig:problem_scenarios}
\vspace{-0.5cm}
\end{figure*}

\subsubsection{FloorPositioning}\label{ssec:floor_positioning} is a simple POMDP benchmark introduced in~\citep{deglurkar2023compositional}. A robot operates in a rectangular environment with internal walls dividing it into upper and lower floors, each with a different goal. The continuous state is the robot position, and the action space comprises motion by $0.05$m in eight compass directions and a \texttt{STAY} action ($|\actSpace|=9$). Actions are deterministic, with collisions leaving the robot stationary. Observations are four-dimensional vectors ($\obsSpace=\mathbb{R}^4$) of distances to the nearest wall or boundary in the cardinal directions, corrupted by zero-mean Gaussian noise with standard deviation $0.01$.

The initial position is sampled such that the robot is uncertain about which floor it occupies. The correct and incorrect goals yield rewards of $100$ and $-100$, respectively, while each non-terminal action incurs $-1$. Episodes terminate upon reaching the correct goal, with discount factor $0.99$.

\vspace{-0.5cm}
\subsubsection{LeggedNavigation}\label{ssec:legged_navigation}

\begin{figure*}[t]
\centering
\includegraphics[width=0.19\textwidth]{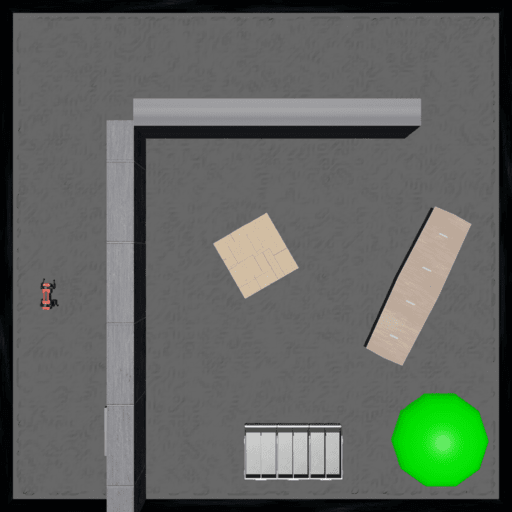}
\includegraphics[width=0.19\textwidth]{images/env_2.png}
\includegraphics[width=0.19\textwidth]{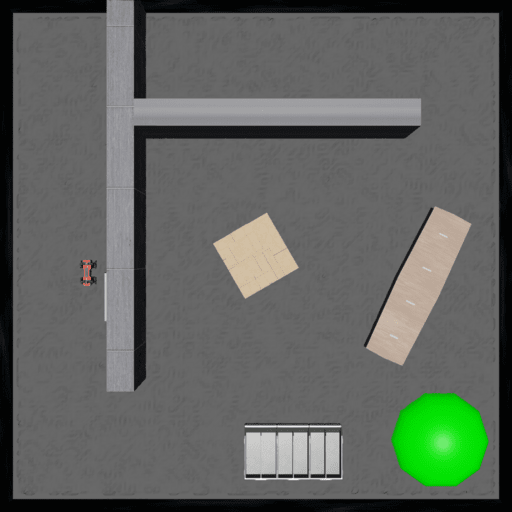}
\includegraphics[width=0.19\textwidth]{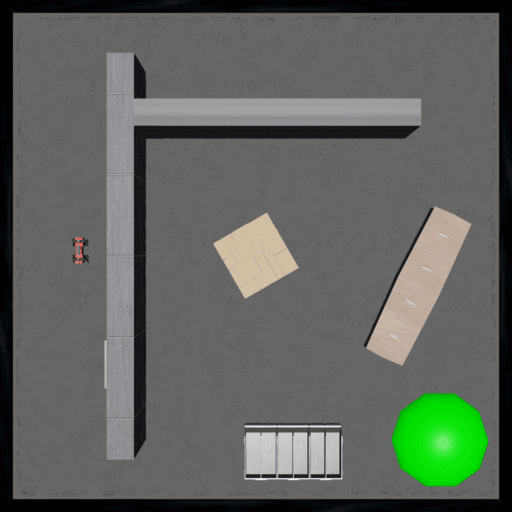}
\includegraphics[width=0.19\textwidth]{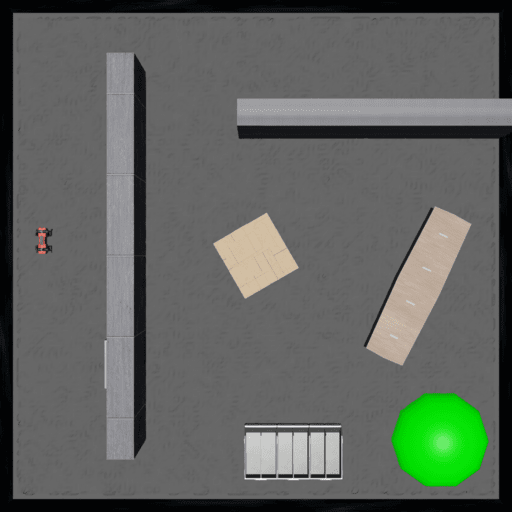}
\caption{Wall configurations used for the \textsc{LeggedNavigation} environment. Different wall configurations induce different traversable routes, which must be inferred from onboard RGB observations. The green region represents the goal.}
\label{fig:legged_navigation_envs}
\vspace{-0.5cm}
\end{figure*}

is a partially observable navigation task, where an\linebreak[4]\mbox{ANYmal-C} quadruped with an onboard RGB camera operates in a cluttered $20\times20$m environment with rough terrain, obstacles, and two internal walls. Starting on the left-hand side of the scene, the robot must navigate to a goal region centered at $(7.5,-7.5)$m (green sphere in \Cref{fig:problem_scenarios}(b)). The two walls can occupy one of five predefined configurations shown in \Cref{fig:legged_navigation_envs}, resulting in different traversable routes. Both the robot's initial position and wall configuration are unknown and must be inferred from RGB observations.

The state comprises the robot position, orientation (represented using a continuous 6D representation~\citep{zhou2019continuity}), linear and angular velocities, joint positions and velocities, and the positions of the two internal walls, resulting in $\stSpace=\mathbb{R}^{39}\times{\texttt{ENV}_1,\ldots,\texttt{ENV}_5}$. Observations are RGB images of resolution $32\times32$, yielding $\obsSpace={0,\ldots,255}^{3\times32\times32}$. The action space consists of six discrete body-velocity commands (forward, slow forward, turn left/right, and forward arcs left/right), which are executed using a pre-trained locomotion controller.

The initial position is sampled uniformly from a $2.3\times5.25$m region on the left-hand side of the environment, while the wall configuration is sampled uniformly from the five layouts in \Cref{fig:legged_navigation_envs}. The initial orientation and joint configuration are assumed known. The objective is to reach a circular goal region of radius $2$m centered at $(7.5,-7.5)$m while avoiding collisions. Reaching the goal yields a reward of $+500$, collisions incur a penalty of $-500$, and each timestep incurs a penalty of $-1$. Episodes terminate upon reaching the goal or collision.

\vspace{-0.5cm}
\subsubsection{TargetFinding}\label{ssec:target_finding}
\textsc{TargetFinding} is a partially observable object-search task evaluating whether POMDP models learned entirely in simulation transfer to physical robot planning. A Unitree Go2 quadruped must locate and reach a target asset (a jerry can) using lidar observations in a $4.9\times3.6$m room containing static obstacles and two cardboard boxes with uncertain positions. The asset is hidden behind one of the boxes, requiring the robot to gather information about the asset and box locations while navigating collision-free.

The state comprises the robot pose, linear and angular velocities, joint positions and velocities, and the asset and box positions, yielding $\stSpace=\mathbb{R}^{45}$. Lidar returns are transformed into the robot frame, restricted to a front-facing field of view, and discretized into $36$ azimuth bins containing the minimum range per sector, normalized by the $2.0$m sensor range and clipped to $[0,1]$, yielding $\obsSpace=\mathbb{R}^{36}$. The same preprocessing is used in simulation and on the physical robot. The action space consists of six discrete body-velocity commands (forward, slow forward, turn left/right, and forward arcs left/right) executed by a pre-trained locomotion controller.

The initial robot position is sampled uniformly from a $1.5\times1.0$m region and its heading within $\pm\pi/8$rad of the nominal orientation. The asset location is sampled uniformly from two candidate locations behind the boxes, while box positions vary by up to $\pm0.5$m longitudinally and $\pm0.25$m laterally around their nominal positions. Reaching the asset yields a reward of $+100$, collisions yield a penalty of $-500$, and each step incurs a cost of $-1$; episodes terminate upon reaching the asset or colliding with a box or environment boundary. All models are trained entirely in simulation and deployed on the physical robot without additional real-world training data.

\vspace{-0.35cm}
\subsection{Experimental Setup}
\vspace{-6pt}

The purpose of our experiments is two-fold. First, we evaluate whether \solverAbbr's learned POMDP models support effective online belief-space planning and generalize to unseen environment configurations. We evaluate this in \textsc{FloorPositioning} and \textsc{LeggedNavigation}, where we compare \solverAbbr against VTS~\citep{deglurkar2023compositional}, DreamerV3~\citep{hafner2025mastering}, and RSAC-D~\citep{ni2022recurrent}. Second, we evaluate whether POMDP models learned from simulation data transfer zero-shot to physical robot planning problems. We investigate this in \textsc{TargetFinding}, where the models learned in simulation are deployed directly for online planning on a physical robot.

For each problem, \solverAbbr learns its task-agnostic POMDP models from $50$K transitions collected offline in simulation using a uniform random policy. In \textsc{LeggedNavigation}, training data is collected only in environments 1 and 2 (\Cref{fig:legged_navigation_envs}) for all methods, while evaluation is performed in all five environments to assess generalization to unseen configurations.

We allocate each method approximately the same total wall-clock offline training budget as \solverAbbr, measured as data collection time plus model or policy training time. Since data collection and training throughput differ between methods and problems, this budget corresponds to different numbers of environment transitions being used by each method. In \textsc{FloorPositioning}, DreamerV3, RSAC-D and VTS use $900$K, $1.8$M and $32$M transitions, respectively; in \textsc{LeggedNavigation}, DreamerV3 and RSAC-D use $600$K transitions, and VTS uses $50$K transitions. We additionally evaluate DreamerV3 and RSAC-D using $2.6$M transitions in both problems, by which point both methods have approximately converged. These variants serve as ablations that assess their performance with substantially more data and computation. Since the official VTS implementation does not support learning transition models, VTS uses the transition sampler learned by \solverAbbr.

For \textsc{LeggedNavigation}, all methods share the same pre-trained locomotion controller, which converts the six high-level body-velocity actions into low-level joint commands. Each high-level action is executed for $2$s. During execution, \solverAbbr and VTS use $1$s of online planning per step in \textsc{FloorPositioning} and \textsc{LeggedNavigation}; \solverAbbr uses $1.5$s in \textsc{TargetFinding}. DreamerV3 and RSAC-D instead execute their learned policies directly, taking $0.01$s/step. We use the R2-Dreamer implementation of DreamerV3\footnote{\url{https://github.com/NM512/r2dreamer}} and the official implementations of RSAC-D for POMDPs\footnote{\url{https://github.com/twni2016/pomdp-baselines/}} and VTS\footnote{\url{https://github.com/michaelhlim/VisualTreeSearch}}. \solverAbbr is implemented in PyTorch~\citep{paszke2019pytorch}. All experiments are carried out on the same machine\footnote{Intel Core i7-13850HX CPU, $32$GB RAM, and NVIDIA RTX 3500 Ada Generation Laptop GPU with $12$GB VRAM.}.

\vspace{-0.35cm}
\subsection{Experimental Results}
\vspace{-6pt}

\begin{table}[ht]
\centering
\vspace{-0.5cm}
\caption{Results on the \textsc{FloorPositioning} problem, averaged over $100$ trials. The best results are highlighted in \textbf{bold}.}
\label{tab:floor_positioning_results}
\footnotesize
\begin{tabular}{lcccc}
\toprule
Metric & \solverAbbr & DreamerV3 & RSAC-D & VTS \\
\midrule
Success rate (\%) $\uparrow$
    & $\bm{100}$ & $\bm{100}$ & $\bm{100}$ & $88$ \\
Avg. total disc. reward $\uparrow$
    & $60.3\pm 1.7$ & $60.7\pm 1.6$ & $\bm{64.4\pm 1.1}$ & $23.5 \pm 11.9$ \\
Training transitions $\downarrow$
    & $\bm{50}\textbf{K}$ & $900$K & $1.8$M & $32$M \\
Data collection time (s) $\downarrow$
    & $\bm{0.71}$ & $19$ & $58$ & $265$ \\
Model/policy training time (s) $\downarrow$
    & $1,665$ & $1,673$ & $1,623$ & $\bm{1,377}$ \\
Total training time & $1,666$ & $1,692$ & $1,681$ & $1,642$ \\
\bottomrule
\end{tabular}
\vspace{-0.5cm}
\end{table}

\begin{table}[t]
\centering
\caption{Results on the \textsc{LeggedNavigation} problem for each environment in \Cref{fig:legged_navigation_envs}, averaged over $50$ trials. The best results are highlighted in \textbf{bold}.}
\label{tab:legged_navigation_performance}
\footnotesize
\setlength{\tabcolsep}{3pt}
\renewcommand{\arraystretch}{0.85}
\begin{tabular}{llcc}
\toprule
Environment & Method \& & Success rate & Avg. total \\
& training transitions & & disc. reward \\
\midrule
1 (trained in) & \solverAbbr ($50$K)             & $94$          & $298.3 \pm 53.6$ \\
  & DreamerV3 (600K)                      & $86$          & $229.7 \pm 78.0$ \\
  & DreamerV3 (2.6M)                      & $88$          & $264.8 \pm 78.7$ \\
  & RSAC-D (600K)                         & $94$          & $342.6 \pm 56.7$ \\
  & RSAC-D (2.6M)                         & $\bm{96}$     & $\bm{357.3 \pm 48.2}$ \\
  & VTS ($50$K)                                 & $14$          & $-136.9 \pm 67.2$ \\
\midrule
2 (trained in) & \solverAbbr ($50$K)             & $\bm{100}$    & $367.5 \pm 5.6$ \\
  & DreamerV3 (600K)                      & $94$          & $321.8 \pm 55.8$ \\
  & DreamerV3 (2.6M)                      & $98$          & $372.5 \pm 32.7$ \\
  & RSAC-D (600K)                         & $94$          & $360.6 \pm 60.1$ \\
  & RSAC-D (2.6M)                         & $98$          & $\bm{388.5 \pm 35.1}$ \\
  & VTS ($50$K)                                   & $42$          & $85.1 \pm 92.9$ \\
\midrule
3 & \solverAbbr ($50$K)                          & $\bm{98}$     & $\bm{395.2 \pm 35.4}$ \\
  & DreamerV3 (600K)                      & $0$           & $-385.0 \pm 38.5$ \\
  & DreamerV3 (2.6M)                      & $0$           & $-376.1 \pm 40.3$ \\
  & RSAC-D (600K)                         & $0$           & $-132.6 \pm 45.0$ \\
  & RSAC-D (2.6M)                         & $0$           & $-432.6 \pm 20.5$ \\
  & VTS ($50$K)                                   & $74$          & $160.9 \pm 97.7$ \\
\midrule
4 & \solverAbbr ($50$K)                          & $\bm{84}$     & $\bm{259.8 \pm 60.3}$ \\
  & DreamerV3 (600K)                      & $46$          & $-108.7 \pm 112.2$ \\
  & DreamerV3 (2.6M)                      & $54$          & $-25.1 \pm 117.8$ \\
  & RSAC-D (600K)                         & $70$          & $131.5 \pm 101.9$ \\
  & RSAC-D (2.6M)                         & $46$          & $-82.5 \pm 124.1$ \\
  & VTS ($50$K)                                  & $32$          & $-102.2 \pm 89.7$ \\
\midrule
5 & \solverAbbr ($50$K)                          & $\bm{88}$     & $\bm{289.8 \pm 62.5}$ \\
  & DreamerV3 (600K)                      & $40$          & $-129.6 \pm 114.4$ \\
  & DreamerV3 (2.6M)                      & $44$          & $-104.0 \pm 118.9$ \\
  & RSAC-D (600K)                         & $48$          & $-59.8 \pm 112.6$ \\
  & RSAC-D (2.6M)                         & $38$          & $-144.3 \pm 123.5$ \\
  & VTS ($50$K)                                  & $42$          & $-43.2 \pm 92.9$ \\
\bottomrule
\end{tabular}
\vspace{-0.5cm}
\end{table}

\begin{table}[htb]
\centering
\caption{Offline training times of all methods in \textsc{LeggedNavigation}.}
\label{tab:legged_navigation_cost}
\footnotesize
\setlength{\tabcolsep}{3pt}
\begin{tabular}{lcccc}
\toprule
Method & Training & Data collection & Model/policy & Total training \\
       & transitions     & time (s)        & training time (s) & time (s) \\
\midrule
\solverAbbr
    & $\bm{50}\textbf{K}$ & $\bm{2,194}$ & $3,758$ & $5,952$\\
DreamerV3 (600K)
    & $600$K & $4,764$ & $\bm{1,248}$ & $6,012$ \\
DreamerV3 (2.6M)
    & $2.6$M & $19,319$ & $5,026$ & $24,345$ \\
RSAC-D (600K)
    & $600$K & $2,841$ & $3,613$ & $6,454$ \\
RSAC-D (2.6M)
    & $2.6$M & $12,186$ & $15,930$ & $28,116$ \\
VTS
    & $\bm{50}\textbf{K}$ & $\bm{2,194}$ & $4,056$ & $6,250$ \\
\bottomrule
\end{tabular}
\end{table}

\noindent
The results in \Cref{tab:floor_positioning_results,tab:legged_navigation_performance,tab:legged_navigation_cost} show that \solverAbbr learns useful task-agnostic POMDP models from substantially less training data than the comparators. In \textsc{FloorPositioning}, \solverAbbr uses at least $18\times$ less training data than the comparators while achieving comparable results. Similarly, in \textsc{LeggedNavigation}, \solverAbbr uses at least $12\times$ less training data while achieving competitive or higher success rates across tested environments. Additionally, on our evaluation hardware and using a batch size of $1,024$, the distilled transition and observation samplers in \textsc{LeggedNavigation} achieved approximately $8.6\times10^5$ and $1.2\times10^6$ samples per second, compared to $8.2\times10^3$ and $1.8\times10^4$ samples per second for the diffusion models, achieving up to $105\times$ higher sampling throughput, while closely matching the diffusion models. On a held-out dataset, the standardized RMSE between samples generated by the diffusion and distilled models is $0.152$ for the transition sampler and $0.274$ for the observation sampler. 

A key advantage of learning task-agnostic POMDP models is their potential to generalize to unseen environments. This is demonstrated in \textsc{LeggedNavigation}, where training data is collected only in environments 1 and 2. Although DreamerV3 and RSAC-D achieve strong performance in the training environments, their success rates drop substantially in the unseen environments 3--5, with a $0\%$ success rate in environment 3. In the training environments, reaching the goal requires the robot to move north and pass through a gap in the upper-left part of the environment. Environment 3 instead requires the robot to move south and pass a lower-left gap, i.e, a qualitatively different navigation strategy from those encountered during training. 

\noindent
Additional training also does not necessarily improve the generalization of task-specific RL policies~\citep{witty2021measuring}. While the $2.6$M-transition RSAC-D policy performs better in the training environments, it performs worse than the $600$K-transition policy in unseen environments 4 and 5. The $2.6$M-transition policy turns toward the upper-left passage earlier and cuts closer around the corner. This produces a shorter route in the training environments, but leaves less clearance in environments 4 and 5, where the passage is narrower, resulting in more collisions.

In contrast, \solverAbbr maintains a success rate of at least $84\%$ across all environments. These results suggest that the learned transition, observation, and likelihood models capture reusable aspects of the environment dynamics and observation process, enabling \solverAbbr to adapt its online policy to previously unseen environment configurations.

\vspace{-0.25cm}
\subsubsection{Results on \textsc{TargetFinding}}
\begin{figure}[htb]
    \centering
\setlength{\tabcolsep}{1pt}
\renewcommand{\arraystretch}{0.8}
    
    \begin{tabular}{cccccc}
        \includegraphics[width=0.155\textwidth]{images/unitree_step_0.png} &
        \includegraphics[width=0.155\textwidth]{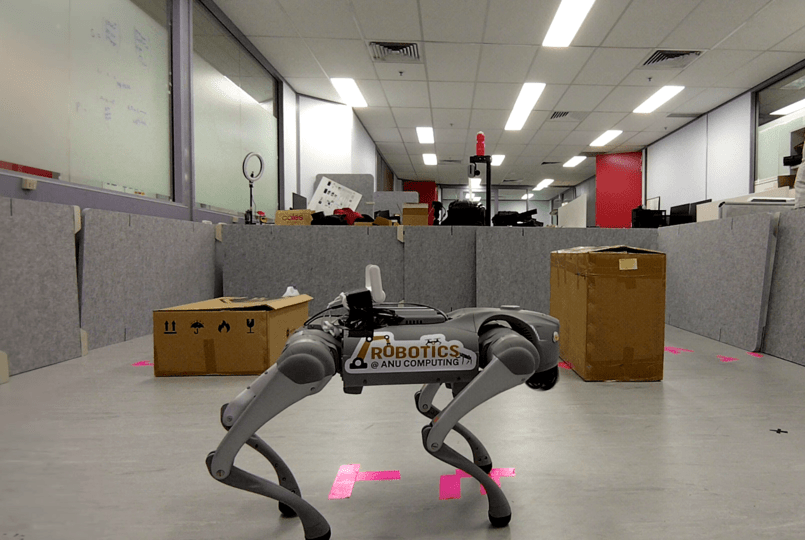} &
        \includegraphics[width=0.155\textwidth]{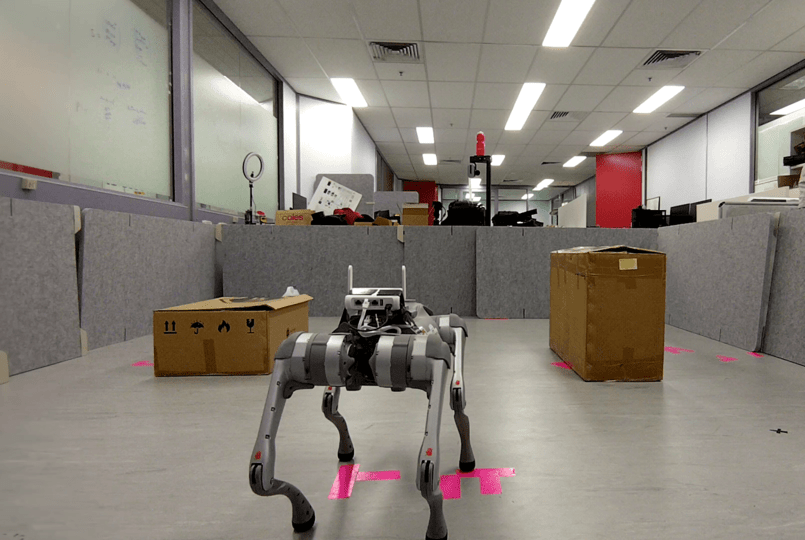} &
        \includegraphics[width=0.155\textwidth]{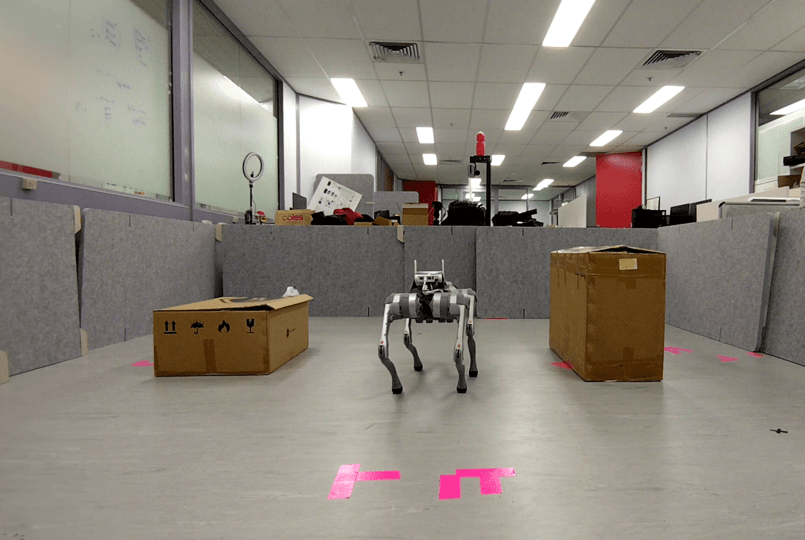} &
        \includegraphics[width=0.155\textwidth]{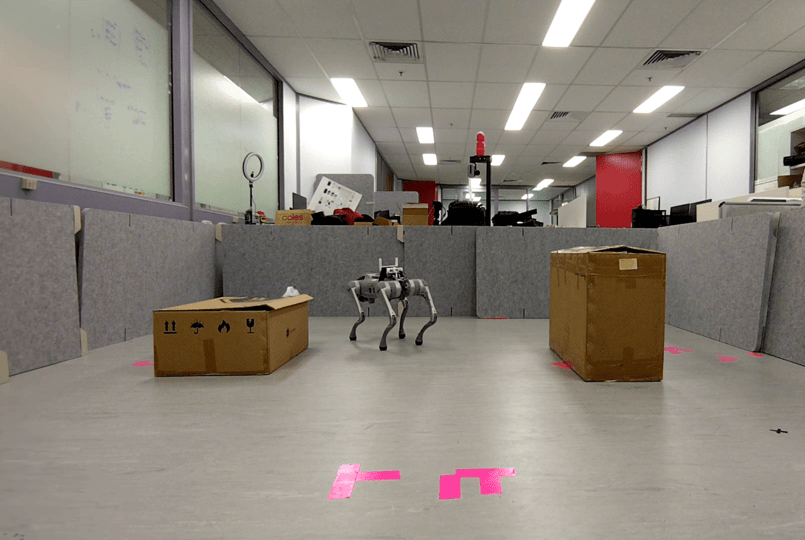} &
        \includegraphics[width=0.155\textwidth]{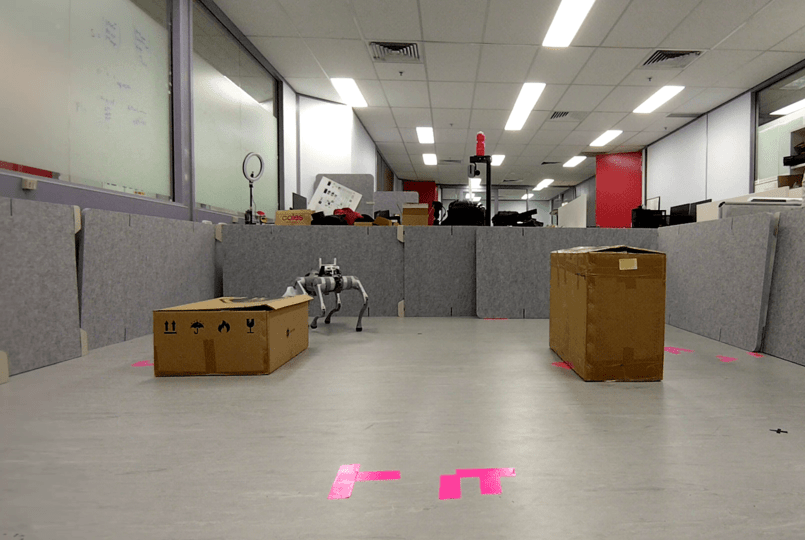} \\
    \end{tabular}

    \vspace{2mm}

    \begin{tabular}{cccccc}
        \includegraphics[width=0.155\textwidth]{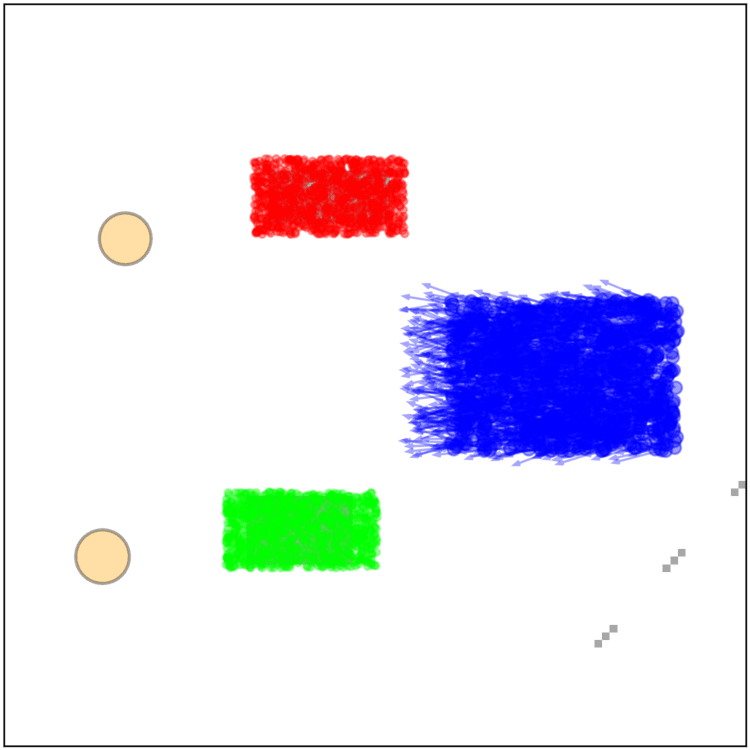} &
        \includegraphics[width=0.155\textwidth]{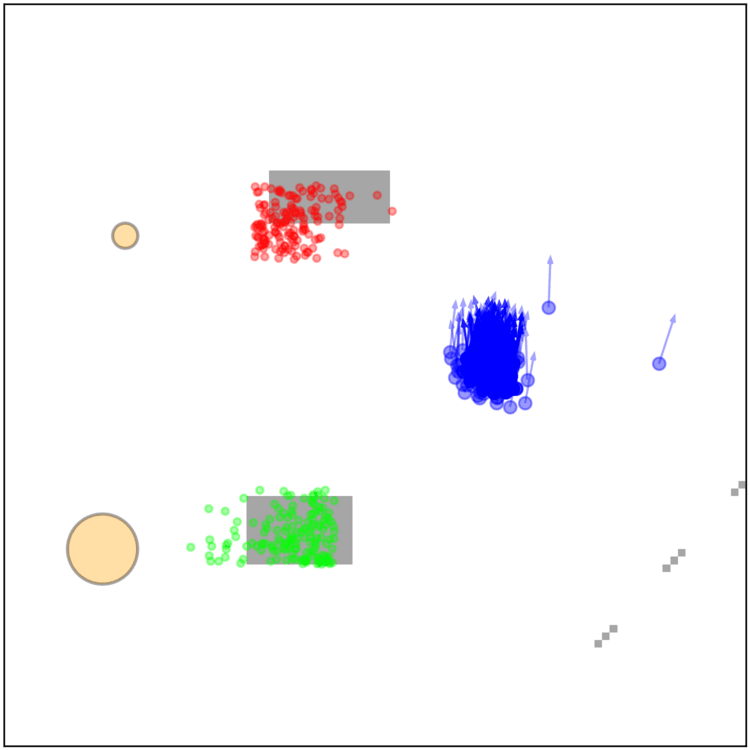} &
        \includegraphics[width=0.155\textwidth]{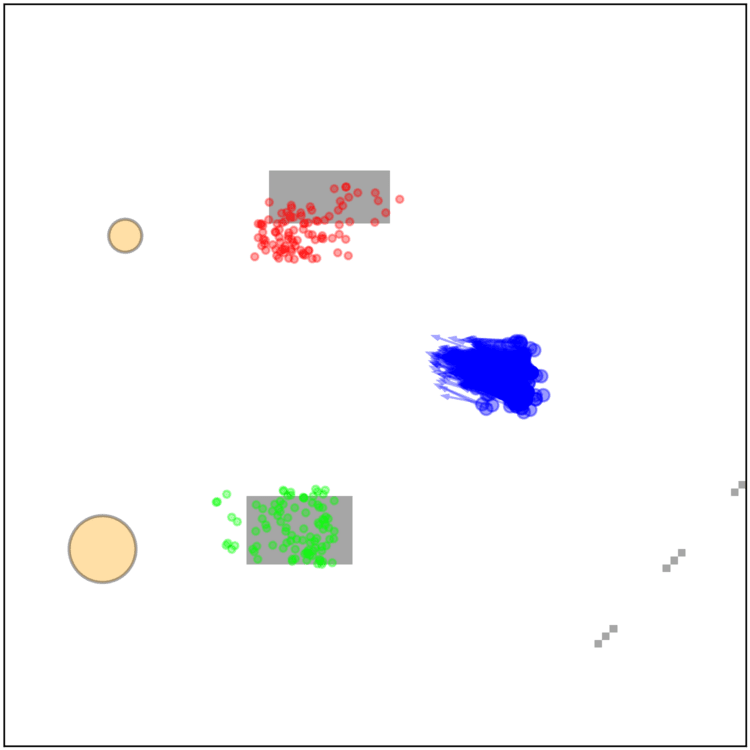} &
        \includegraphics[width=0.155\textwidth]{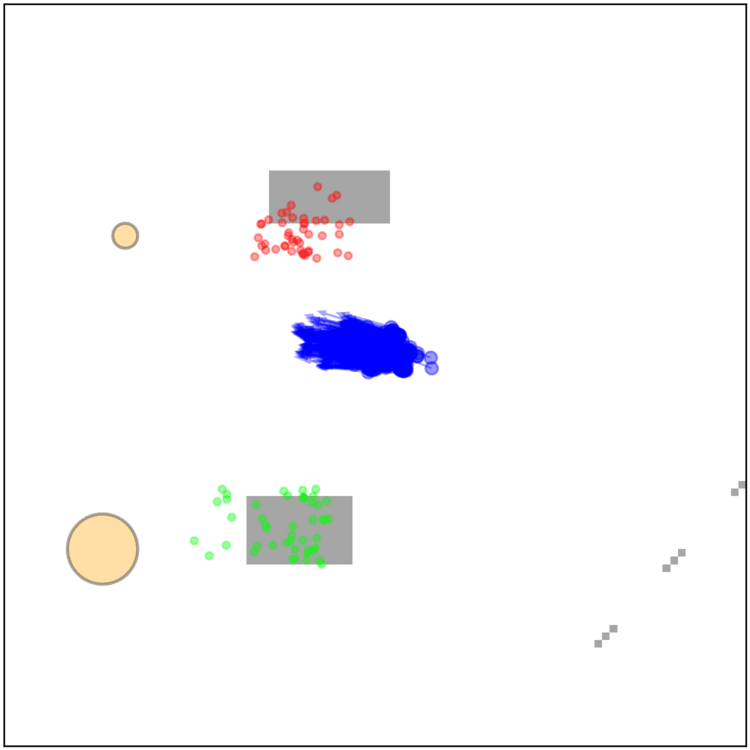} &
        \includegraphics[width=0.155\textwidth]{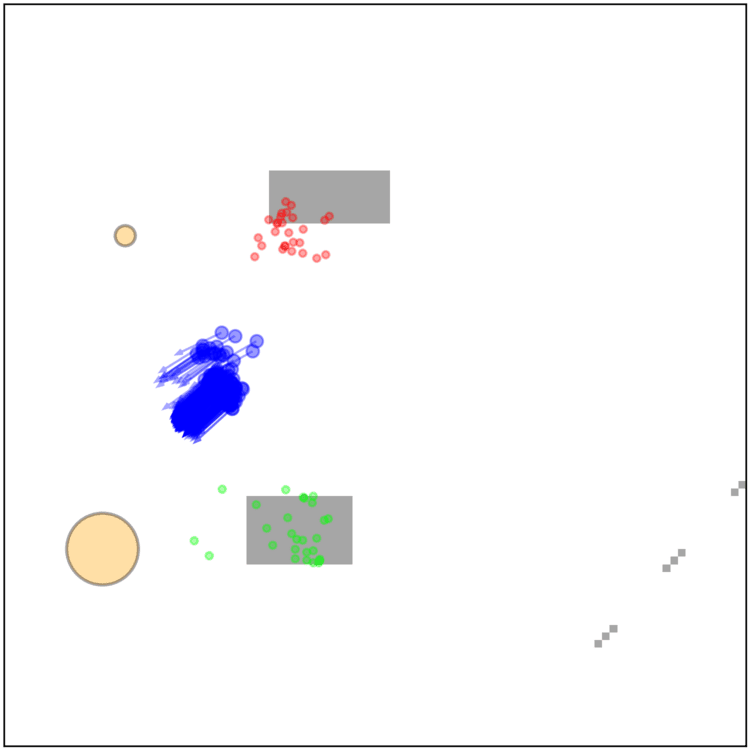} &
        \includegraphics[width=0.155\textwidth]{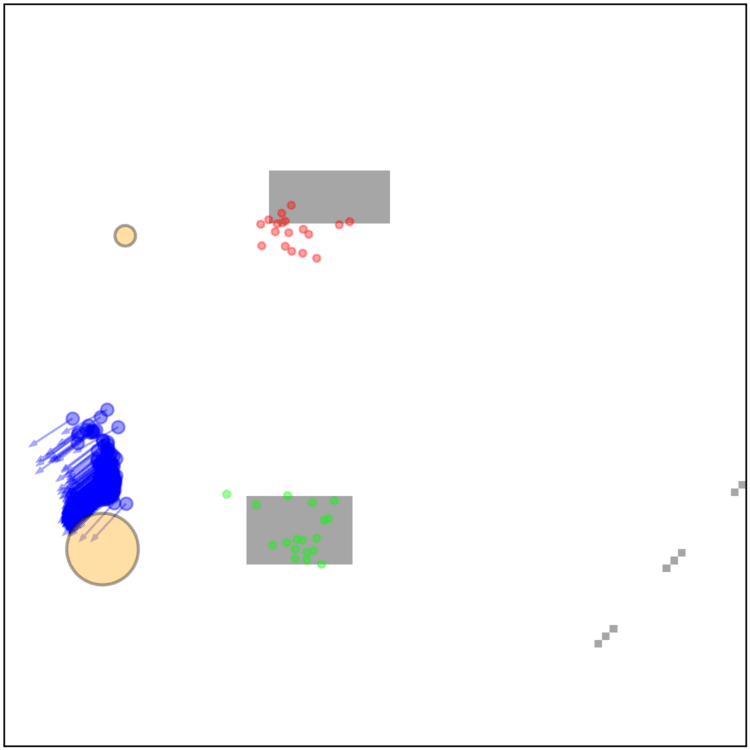} \\
        \small Step 0 &
        \small Step 1 &
        \small Step 2 &
        \small Step 3 &
        \small Step 4 &
        \small Step 5 \\
    \end{tabular}
    \caption{
Real-world execution of the Unitree Go2 quadruped and corresponding belief evolution.
The top row shows images caputed during execution. The bottom row shows beliefs over robot pose (blue particles), asset location (orange circles), and movable obstacle locations (green and red particles). As observations are incorporated, the belief progressively concentrates around the true environment state.
}    
    \label{fig:unitree_belief_evolution}
    \vspace{-0.5cm}
\end{figure}
To see whether POMDP models learned in simulation transfer to a real robot, we deployed \solverAbbr on \textsc{TargetFinding}. Similar to the previous benchmarks, the transition, observation, and likelihood models were trained using $50$K transitions collected in simulation and were transferred to the real robot without further training. Across $10$ real-world test runs, including five runs for each possible target location, \solverAbbr successfully completed the task in every trial. \Cref{fig:unitree_belief_evolution} shows an example run of the problem, while the accompanying video shows a run when the robot initially identifies a wrong target location, explores and finally finds the target. To introduce additional variability, the positions of the cardboard obstacles were manually changed between runs. These results suggest that the learned transition, observation, and likelihood models are sufficiently robust to transfer from simulation to a physical robot.

\vspace{-0.25cm}
\section{Conclusion}\label{sec:conclusions}
\vspace{-0.25cm}
We present \solverAbbr, a framework for approximate POMDP solving without a priori models. \solverAbbr consists of two components: a 
task-agnostic POMDP model learning designed for online planning, and a scalable fully vectorized POMDP planner, VOPP. The model learning  component combines diffusion-based generative modeling with model distillation to learn task-agnostic transition and observation samplers and observation-likelihood models that support both particle-based belief updates and online belief-space planning. \solverAbbr enables efficient belief-space planning while retaining the ability to represent complex transition and high-dimensional observation functions. Experimental results on three benchmark problems demonstrate that \solverAbbr can learn efficient POMDP models from relatively small amounts of training data, generalize to unseen environment configurations, and transfer successfully from simulation to real-world planning problems without additional training. These results suggest that combining reusable learned POMDP models with scalable online planning is a promising direction for robotic sequential decision making under uncertainty.

\vspace{-12pt}
\section*{Acknowledgements}
\vspace{-9pt}
{\small
This work was supported by the ARC Research Hub in Intelligent Robotic Systems for Real-Time Asset Management (IH210100030), in collaboration with the University of Sydney and Nexxis Technology.
}

%
%
\renewcommand{\bibname}{References\vspace{-2em}}
\begingroup
\let\oldclearpage\clearpage
\let\oldcleardoublepage\cleardoublepage
\renewcommand{\clearpage}{}
\renewcommand{\cleardoublepage}{}
\bibliographystyle{spmpsci}
\bibliography{bibliography}
\let\clearpage\oldclearpage
\let\cleardoublepage\oldcleardoublepage
\endgroup

\end{document}